\documentclass{ifacconf}

\usepackage{graphicx}      
\usepackage{natbib}        

\usepackage{svg}
\usepackage{subcaption}
\usepackage{multirow}
\usepackage{booktabs}
\usepackage{caption}
\usepackage{graphicx}
\usepackage{adjustbox}
\usepackage{svg}
\svgsetup{
  inkscapepath=svg-inkscape, 
  inkscapearea=page,
  inkscapelatex=false,       
  inkscapeopt=-z -D --export-text-to-path, 
}
\usepackage{siunitx}

\usepackage{amsmath} 
\usepackage{amssymb}  
\usepackage{acro}
\usepackage{bm}

\acsetup{patch/maketitle=false}
\DeclareAcronym{AGV}{
    short = AGV,
    long = Autonomous Ground Vehicle
}

\DeclareAcronym{TD3}{
    short = TD3, 
    long = Twin Delayed Deep Deterministic Policy Gradient,
}

\DeclareAcronym{DRL}{
    short = DRL,
    long = Deep Reinforcement Learning
}
\DeclareAcronym{SFM}{
    short = SFM,
    long = Social Force Model
}

\DeclareAcronym{ORCA}{
    short = ORCA,
    long = Optimal Reciprocal Collision Avoidance
}

\DeclareAcronym{POMDP}{
    short = POMDP,
    long = Partially Observable Markov Decision Process
}

\DeclareAcronym{LSTM}{
    short = LSTM,
    long = Long Short-Term Memory
}

\DeclareAcronym{PPO}{
    short = PPO,
    long = Proximal Policy Optimization
}

\DeclareAcronym{ODV-PPO}{
    short = ODV-PPO,
    long = Observation-Dependent Variance PPO
}

\DeclareAcronym{SAC}{
    short = SAC,
    long = Soft Actor-Critic
}

\DeclareAcronym{NN}{
    short = NN,
    long = Neural Network
}

\DeclareAcronym{OOD}{
    short = OOD,
    long = Out-Of-Distribution
}

\DeclareAcronym{MC-Dropout}{
    short = MC-dropout,
    long = Monte-Carlo dropout
}

\DeclareAcronym{MPC}{
    short = MPC,
    long = Model-Predictive Controller

}
\DeclareAcronym{ODV}{
    short = ODV,
    long = Observation-Dependent Variance
}
\DeclareAcronym{MSE}{
    short = MSE,
    long = Mean Squared Error
}
\DeclareAcronym{DQN}{
    short = DQN,
    long = Double Q-Network
}

\DeclareAcronym{PCN}{
    short= PCN,
    long = Probability of Collision Network,
}

\DeclareAcronym{POC}{
    short= POC,
    long = Probability of Collision,
}

\DeclareAcronym{SCA}{
    short= SCA,
    long = Social Collision Avoidance,
}

\DeclareAcronym{SA}{
    short= SA,
    long = Socially Aware
}

\DeclareAcronym{SI}{
    short= SI,
    long = Socially Integrated,
}

\DeclareAcronym{ANN}{
    short= ANN,
    long= Artificial Neural Network,
    short-plural = s ,
    long-plural = s ,
}

\DeclareAcronym{SNN}{
    short= SNN,
    long= Spiking Neural Network,
    short-plural = s ,
    long-plural = s ,
}

\DeclareAcronym{SAN}{
    short= SAN,
    long= Spiking Actor Network,
    short-plural = s ,
    long-plural = s ,
}

\DeclareAcronym{SFE}{
    short= SFE,
    long= Spiking Feature Extractor,
    short-plural = s ,
    long-plural = s ,
}

\DeclareAcronym{GRF}{
    short= GRF,
    long= Gaussian Receptive Field,
    short-plural = s ,
    long-plural = s ,
}

\DeclareAcronym{RNN}{
    short= RNN,
    long= Recurrent Neural Network,
    short-plural = s ,
    long-plural = s ,
}

\DeclareAcronym{ALIF}{
    short= ALIF,
    long= Adaptive Leaky Integrate and Fire,
    short-plural = s ,
    long-plural = s ,
}

\DeclareAcronym{CUBA}{
    short= CUBA,
    long= Current Based,
    short-plural = s ,
    long-plural = s ,
}

\DeclareAcronym{SD}{
    short= SD,
    long= Sigma-Delta,
    short-plural = s ,
    long-plural = s ,
}

\DeclareAcronym{SYNOPS}{
    short= SYNOPS,
    long= Synaptic Operations,
}

\DeclareAcronym{PV}{
    short= PV,
    long= proxemic violation,
    short-plural = s ,
    long-plural = s ,
}

\DeclareAcronym{CPU}{
    short= CPU,
    long= Central Processing Unit,
    short-plural = s ,
    long-plural = s ,
}

\DeclareAcronym{GPU}{
    short= GPU,
    long= Graphics Processing Unit,
    short-plural = s ,
    long-plural = s ,
}

\DeclareAcronym{RVO}{
    short= RVO,
    long= Reciprocal Velocity Obstacle,
    short-plural = s ,
    long-plural = s ,
}

\DeclareAcronym{HRVO}{
    short= HRVO,
    long= Hybrid Reciprocal Velocity Obstacle,
    short-plural = s ,
    long-plural = s ,
}

\DeclareAcronym{SA-CADRL}{
    short= SA-CADRL,
    long= Social Aware Collision Avoidance with DRL,
    short-plural = s ,
    long-plural = s ,
}

\DeclareAcronym{GRU}{
    short= GRU,
    long= Gated Recurrent Unit,
    short-plural = s ,
    long-plural = s ,
}

\DeclareAcronym{CASRL}{
    short= CASRL,
    long= Collision Avoidance Spiking Reinforcement Learning,
    short-plural = s ,
    long-plural = s ,
}

\DeclareAcronym{SDDPG}{
    short= SDDPG,
    long= Spiking Deep Deterministic Policy Gradient,
    short-plural = s ,
    long-plural = s ,
}

\DeclareAcronym{LIF}{
    short= LIF,
    long= Leaky Integrate and Fire,
}

\DeclareAcronym{TO}{
    short = TO, 
    long = timeout
}

\DeclareAcronym{DR}{
    short = DR, 
    long = distance ratio
}

\begin{document}


\begin{frontmatter}
\vspace{-3.5em}\footnotesize This work has been submitted to the IFAC for possible publication.\\
Copyright may be transferred without notice, after which this version will no longer be accessible.\\
\title{SINRL: Socially Integrated Navigation with Reinforcement Learning using Spiking Neural Networks} 


\author[First]{Florian Tretter} 
\author[First]{Daniel Flögel} 
\author[First]{Alexandru Vasilache}
\author[First]{Max Grobbel}
\author[Second]{Jürgen Becker}
\author[Second]{Sören Hohmann}

\address[First]{FZI Research Center for Information Technology, \\76131 Karlsruhe, Germany (e-mail: floegel@fzi.de)}
\address[Second]{Karlsruhe Institute of Technology, 76131 Karlsruhe, Germany (e-mail: soeren.hohmann@kit.edu)}

\begin{abstract}                
Integrating autonomous mobile robots into human environments requires human-like decision-making and energy-efficient, event-based computation.
Despite progress, neuromorphic methods are rarely applied to \ac{DRL} navigation approaches due to unstable training.
We address this gap with a hybrid socially integrated \ac{DRL} actor-critic approach that combines \acp{SNN} in the actor with \acp{ANN} in the critic and a neuromorphic feature extractor to capture temporal crowd dynamics and human-robot interactions. 
Our approach enhances social navigation performance and reduces estimated energy consumption by approximately 1.69 orders of magnitude.

\end{abstract}


\end{frontmatter}

\section{Introduction}
Intelligent autonomous mobile robots are increasingly integrated in crowded environments to assist humans. 
Their applications range from service robots in hospitals and airports to last-mile delivery systems in urban areas. 
As these robots become more pervasive, recent research has begun to explore neuromorphic computing to achieve more energy-efficient systems \citep{xu_mapless_navigation}, as well as biological-inspired approaches \citep{shim_biological_inspired_coll} with event-based computation.
Despite the progress in social navigation, two fundamental challenges remain for autonomous robots in crowded environments.
First, fully leveraging the benefits of neuromorphic computing requires end-to-end architectures in which both perception and decision-making are realized on neuromorphic hardware.
Second, because autonomous mobile robots navigate in human-centered environments, social navigation approaches must demonstrate socially compliant behavior to ensure safe and acceptable interactions with humans.

Social navigation is a subfield of local motion planning, focusing on solving navigation tasks in highly dynamic crowded environments.
\cite{MavrogiannisCoreChallengesofSocialRobotNavigati2021} divides social navigation approaches into coupled and decoupled approaches.
Decoupled approaches first predict human motions and subsequently plan a collision-free trajectory, which can lead to the freezing robot problem as described by \cite{TrautmanUnfreezingtherobotNavigationindens2010}.
Coupled approaches consider the future crowd evolution as a joint sequential decision making and either assume a specific structure (e.g. game theoretic in \cite{SamaviSICNav2025}, forces based in \cite{HelbingSocialforcemodelforpedestriandynami1995} or reciprocal decision making in \ac{ORCA} from \cite{vandenBergReciprocalnBodyCollisionAvoidance2011}), or assume insights about the principle decision making problem as in \ac{DRL} approaches first proposed by \cite{ChenDecentralizednoncommunicatingmultiage2017}.
\ac{DRL} based approaches are further categorized based on the robots exhibited social behavior \citep{FloegelSociallyIntegratedNavigation2024}.
Social collision avoidance, with a lack of social aspects, socially aware navigation with a predefined social behavior, and socially integrated navigation, derived from sociological definitions, where the robot's social behavior is adaptive to individual human behavior and emerges through interaction.

Although recent work such as \cite{SaravananExploringSpikingNeuralNetworksinSingleandMultiagentRLMethods2021} demonstrates that \acp{SNN} can improve policy search in continuous control \ac{DRL} tasks, a proper hyperparameter tuning is crucial for stable training. 
Furthermore, research on applying \acp{SNN} to social navigation remains sparse, limited mostly to social collision avoidance with hybrid \ac{DRL} \ac{ANN}-\ac{SNN} approaches, while socially aware or socially integrated navigation is entirely unexplored.
Therefore, we propose a socially integrated hybrid \ac{DRL} approach, which enables an end-to-end neuromorphic path for policy inference, resulting in improved social navigation performance and reduced energy consumption.

The main contribution of this work is a novel socially integrated hybrid \ac{DRL} navigation approach that leverages the sparsity of \acp{SNN} for energy reductions and event-based computation of human-robot interactions.
We first design a hybrid actor-critic \ac{DRL} architecture with \acp{SNN} in the actor and the training stability of hyperparameter-insensitive \acp{ANN} in the critic network.
Subsequently, a neuromorphic feature extractor with an inherent temporal dimension is used to cope with observing a variable number of surrounding agents and capture event-based crowd-robot interactions. 
Finally, we benchmark our approach against state-of-the-art socially integrated and socially aware navigation approaches, and compare the estimated energy consumption of the trained models on various conventional and neuromorphic hardware platforms.

\section{Preliminaries}
A dynamic object in the environment is generally referred to as an agent, whether it is a robot or a human, and a policy determines its behavior.
Variables referring to the robot are indexed with $\square^0$, and humans with $\square^i$, where $i \in 1, \cdots, N-1$. 
A scalar value is denoted by $\square$, a vector $\bm{\square}$ is bold, and a matrix is capital bold.

\subsection{Problem Formulation}

The navigation task of one robot toward a goal in an environment with $N-1$ humans is a sequential decision-making problem that can be modeled as an \ac{POMDP} and solved with a \ac{DRL} framework \citep{ChenDecentralizednoncommunicatingmultiage2017}. 
The \ac{POMDP} is described with a 8-tuple $(\mathcal{S}, \mathcal{A}, \mathcal{T}, \mathcal{O}, \Omega, \mathcal{T}_0, R, \gamma)$.
We assume the state space $\mathcal{S} \in \mathbb{R}$ and action space $\mathcal{A} \in \mathbb{R}$ as continuous.
The transition function $\mathcal{T} : \mathcal{S} \times \mathcal{A} \times \mathcal{S} \rightarrow [0,1]$ describes the probability transitioning from state $\bm{s}_{t} \in \mathcal{S}$ to state $\bm{s}_{t+1} \in \mathcal{S}$ for the given action $\bm{a}_t \in \mathcal{A}$. 
With each transition, an observation $\bm{o}_t \in \mathcal{O}$ and a reward $R : \mathcal{S} \times \mathcal{A} \rightarrow \mathbb{R}$ is returned by the environment. 
The observation $\bm{o}_t$ is returned with probability $\Omega(\bm{o}_t|\bm{s}_t)$ depending on the sensors.
The initial state distribution is denoted by $\mathcal{T}_0$ and $\gamma \in [0,1)$ is the discount factor.
We assume circular shaped agents and each agent, robot and humans, is completely described with a state $\bm{s}_{t}^{i} = [\bm{s}_{t}^{i,\mathrm{o}} , \bm{s}_{t}^{i,\mathrm{h}}]$ with $i \in 0, \cdots N-1$ at any given time $t$.
The observable part $\bm{s}_{t}^{i,\mathrm{o}} = [\bm{p}, \bm{v}, r]$ is composed of position $\bm{p} $, velocity $\bm{v}$, and radius $r$.
The unobservable, hidden part, $\bm{s}_{t}^{i,\mathrm{h}} = [\bm{p}_{\mathrm{g}}, v_{\mathrm{pref}}, \psi_{\mathrm{pref}}, r_{\mathrm{prox}}]$ is composed of goal position $\bm{p}_{\mathrm{g}}$, preferred velocity $v_{\mathrm{pref}}$, preferred orientation $\psi_{\mathrm{pref}}$, and a proxemic radius $r_{\mathrm{prox}}$ according to the proxemic theory of \cite{HallThehiddendimension1990}.
The world state $\bm{s}_t = [\bm{s}_{t}^0, \cdots, \bm{s}_{t}^{N-1}]$ represents the environment at time $t$.
One episode's trajectory $\tau$ is the sequence of states, observations, actions, and rewards over the time interval $t\in [0,T]$.
The return of one episode $\mathcal{R}(\tau) = \sum_{t=0}^T \gamma^t R_t$ is the accumulated and discounted reward $R_t$, and the central objective is to learn the optimal robot policy $\pi^*$ which maximizes the expected return:
\begin{align}
    \mathcal{T}(\tau|\pi) &= \mathcal{T}_0 \prod_{t=0}^T \mathcal{T}(\bm{s}_{t+1}|\bm{s}_t, \bm{a}_t) \pi(\bm{a}_t|\bm{o}_t)\Omega(\bm{o}_t|\bm{s}_t)\text{,}\\
    \underset{\tau \sim \pi}{\mathbb{E}} [\mathcal{R}(\tau)] &= \int_\tau \mathcal{T}(\tau|\pi)\mathcal{R}(\tau) \text{,}\\
    \pi^*(\bm{a}|\bm{o}) &=  \arg \underset{\pi}{\max} \underset{\tau \sim \pi}{\mathbb{E}} [\mathcal{R}(\tau)]\text{.}
\end{align}
Considering a stochastic environment, $\mathcal{T}(\tau|\pi)$ is the probability of a trajectory $\tau$ starting in $\bm{s}_0$ with $\mathcal{T}_0$.

\subsection{Spiking Neural Networks}
\acsp{SNN} represent the third generation of neural networks and are biologically inspired \citep{shim_biological_inspired_coll}.
The key difference to \acsp{ANN} is the introduction of a temporal dimension within the single neuron dynamics.
The communication of the neurons relies on binary or graded spikes, which are summed up in the dendrite of each neuron.
The neuron then processes the accumulated spikes according to its neuron dynamics and sends out spikes.
Numerous neuron models exist, each with distinct dynamics that emphasize biological plausibility, such as the Izhikevich neuron \citep{IzhikevichResonate-and-fireneurons2001}, computational efficiency, as seen in the \ac{LIF} neuron, or sparsity, as in the \ac{SD} neuron.
In this work, we investigate the \ac{CUBA} and \ac{SD} neuron model.

The \ac{CUBA} neuron implements a current-based integration mechanism and is a computationally efficient simplified version of the \ac{LIF}. The membrane potential $V(t)$ evolves according to
\begin{equation}
\tau_\mathrm{m} \frac{dV(t)}{dt} = - (V(t) - E_\mathrm{L}) + R_\mathrm{m} I_{\mathrm{syn}}(t),
\end{equation}
with membrane time constant $\tau_\mathrm{m}$, resting potential $E_\mathrm{L}$, and total synaptic input current $I_{\mathrm{syn}}(t)$.
A spike is emitted when $V(t)$ exceeds a threshold $V_{\mathrm{th}}$, followed by a reset.
This formulation assumes current-based synaptic transmission independent of the membrane potential, resulting in linear and additive input dynamics.
The reduced computational complexity of the \ac{CUBA} model makes it well-suited for reinforcement learning and control applications requiring efficient large-scale spiking architectures.

In contrast, the \ac{SD} neuron encodes information as the temporal error between the input and its reconstructed output signal, following the principles of sigma-delta modulation.
Its internal state $u(t)$ evolves as
\begin{equation}
\tau_\mathrm{m} \frac{du(t)}{dt} = I(t) - \hat{I}(t),
\end{equation}
where $I(t)$ represents the input current and $\hat{I}(t)$ denotes the reconstructed feedback current derived from emitted spikes.
A spike is generated when the integrated error exceeds a threshold, and the neuron resets its internal state accordingly.
This mechanism enables \ac{SD} neurons to operate as adaptive, event-driven encoders with inherent temporal sparsity, allowing for efficient communication and energy usage.
Compared to the CUBA neuron, the \ac{SD} neuron model captures differential input dynamics, making it particularly beneficial for control tasks requiring responsiveness to rapid state changes.

\section{Related Work}

Recent progress in energy-efficient neuromorphic hardware has sparked growing interest in applying \acp{SNN} to robotic systems, demonstrating strong potential within \ac{DRL} \citep{Jiang_Neuro_Planner}.
Hybrid approaches for single-agent continuous control, such as the work by \cite{tang_deep_2021}, achieved a 140× reduction in power consumption on Intel’s Loihi neuromorphic processor compared to an \ac{ANN} baseline running on the Jetson TX2.
To address \ac{POMDP} scenarios, \cite{cheng_spiking_2023} extended a spiking \ac{TD3} agent with an \ac{LSTM}-based feature extractor, resulting in 50–80\% lower deployment energy relative to ANN counterparts on PyBullet robotics benchmarks.
In the context of multi-agent continuous control, \cite{SaravananExploringSpikingNeuralNetworksinSingleandMultiagentRLMethods2021} reported faster training convergence and higher returns compared to ANN-based implementations.
These studies highlight that neuromorphic deployment is practical and highly efficient, and current efforts are mainly focused on continuous control tasks.

Recent approaches in social navigation aim to exhibit human-like social behavior and decision-making. 
While state-of-the-art MPC-based approaches, such as \cite{SamaviSICNav2025}, jointly optimize robots and the crowd's motion in a bi-level structure, they rely on specific assumptions about the human motion model in the optimization problem, which do not consider social aspects, such as proxemics. 
In contrast, socially integrated \ac{DRL} navigation approaches, as proposed by \cite{FloegelSociallyIntegratedNavigation2024}, do not assume a specific human motion model; they learn a navigation policy that acts according to the actions of others and treats humans individually. 
With this formulation, these approaches fulfill the sociological understanding of social acting and socially acceptable behavior. 
For a more detailed description, the reader is referred to \cite{FloegelSociallyIntegratedNavigation2024} for socially integrated navigation, to \cite{RiosMartinezFromProxemicsTheorytoSociallyAware2015} for proxemic theory, and to \cite{MavrogiannisCoreChallengesofSocialRobotNavigati2021} for core challenges in social navigation.


In \ac{DRL} social navigation architectures, a feature extractor typically precedes the actor and critic to capture crowd dynamics and handle a varying number of observed agents.
In \ac{ANN}-based methods, this extractor is often a \ac{RNN} (e.g., an \ac{LSTM} as proposed by \cite{everett_collision_2021}) to handle an arbitrary number of agents by stacking their observations into a sequence fed to the network.
The \ac{RNN} accumulates information in its hidden state, producing a fixed-size feature vector used by the actor and critic.
However, this sequential \ac{RNN} extractor is order-sensitive, scales poorly, and fails to capture higher-order agent interactions \citep{ChanganChenCrowdRobotInteraction:CrowdawareRob2019}.

\ac{SNN}-based approaches are often employed in hybrid configurations.
\cite{tang_reinforcement_2020} advanced this direction by jointly training spiking and non-spiking network components with their \ac{SDDPG} algorithm.
When deployed on Intel’s Loihi neuromorphic processor, their method achieved higher navigation success rates and reduced energy consumption by a factor of $75$ compared to baseline \ac{DRL} policy on an NVIDIA Jetson TX2.
\cite{yang_spiking_2023} extended this hybrid method to social collision avoidance by introducing a spiking \ac{GRU} to provide memory for multi-agent navigation.
Their multi-critic architecture outperformed Tang’s \ac{SDDPG} in previously unseen scenarios and produced more efficient trajectories.
A related line of work was proposed by \cite{chengjun_zhang1_casrl_2024}, who introduced \ac{CASRL}, a hybrid framework that integrates a spiking transformer encoder with attention-based temporal fusion.
Yielding improved capturing of human interaction patterns.
\ac{CASRL} outperformed established baselines—including \ac{ORCA}, \ac{SA-CADRL}, and \ac{SDDPG}—in multi-agent settings while reducing energy consumption to $57\%$ of the \ac{ANN} counterpart.

With the promising results of hybrid \ac{DRL} on social collision avoidance tasks, the research question arises whether \ac{SNN} based \ac{DRL} policies can capture the nuanced social dynamics of human–robot interaction for socially integrated navigation tasks while simultaneously reducing the energy consumption. 

\begin{figure}[tb]
    \centering
    \includegraphics[width=0.9\linewidth]{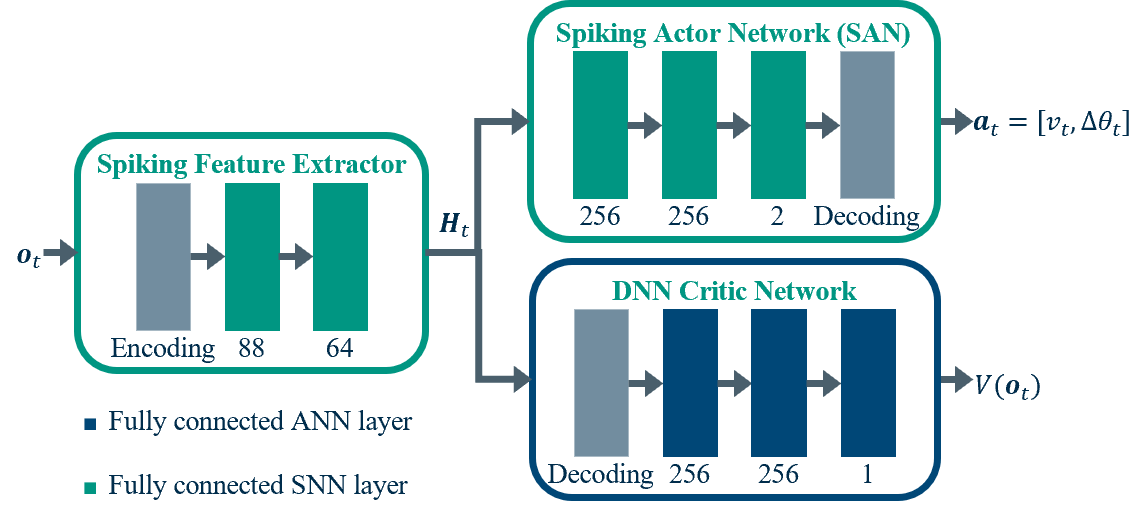}
    \caption{Architecture composition with a temporal spiking feature output $\mathbf{H}_t$ achieving a end-to-end spiking path.}
    \label{fig:shared-fe-actor-critic-arcitecture}
\end{figure}
\section{Approach}
We propose a hybrid \ac{DRL} framework that utilizes the \ac{PPO} algorithm \citep{SchulmanProximalPolicyOptimizationAlgorithms2017} with an \ac{SNN}-based actor, an \ac{ANN}-based critic, and a preceding \ac{SFE} as depicted in Fig.~\ref{fig:shared-fe-actor-critic-arcitecture}. 
The \ac{SFE} leverages the inherent temporal dynamics of \acsp{SNN} to more effectively capture human-robot interactions and learns an embedding of the crowd-robot interaction.
This design combines the advantages of sparse and event-driven inference from the \ac{SNN} for potentially lower energy consumption with the stable gradient-based learning during training from the \ac{ANN}.

\subsection{Hybrid Approach}
We follow the framework in \cite{FloegelSociallyIntegratedNavigation2024} to train a \ac{SI} navigation policy and use the same observation (Appx.~\ref{appendix:obs}) and reward system (Appx.~\ref{appendix:rew}).
The policy is trained from scratch, respects proxemic and velocity social norms, and exhibits social behavior that is adaptive to individual human preferences.
The action $\bm{a}_t$ of the policy is a velocity $v_t$ and a delta heading $\Delta \theta_t$ command. 
The hybrid actor-critic approach integrates a \ac{SAN} with a conventional \ac{ANN} critic (Fig.~\ref{fig:shared-fe-actor-critic-arcitecture}).
The shared \ac{SFE} allows multi-agent observations by transforming the observation into a temporally encoded feature tensor.
In contrast to recurrent structures that rely on artificial temporal dependencies, the \ac{SFE} represents sequential dependencies through the intrinsic temporal dynamics of spiking neurons, without requiring explicit recurrent structures.
The feature tensor $\mathbf{H}_t$ serves as a direct input to the actor, whereas the critic relies on a decoder that transforms the spikes back into the floating-point domain.
The critic outputs the value function $V(\mathbf{o}_t)$, evaluating the \ac{SAN} and thus serving as a learning signal. 

During training, the \ac{SFE}, \ac{SAN}, and the critic are jointly optimized using \ac{PPO}.
The \ac{ANN}-based critic operates in the floating-point domain to ensure stable learning, while the feature extractor and actor remain in the spiking domain, enabling compatibility with neuromorphic hardware.
During inference, only the spiking pathway from the \ac{SFE} to the \ac{SAN} is required, yielding a fully neuromorphic execution pipeline from observation to action.
This is particularly advantageous for deployment on neuromorphic platforms and in combination with neuromorphic perception systems.
This hybrid approach decouples training efficiency from inference efficiency, allowing learning to benefit from the stability of conventional ANN training while deployment exploits the event-driven, low-power characteristics of neuromorphic computation.

\begin{figure}[tb]
    \centering
    \begin{subfigure}[b]{0.48\linewidth}
        \centering
        \includegraphics[width=0.95\linewidth]{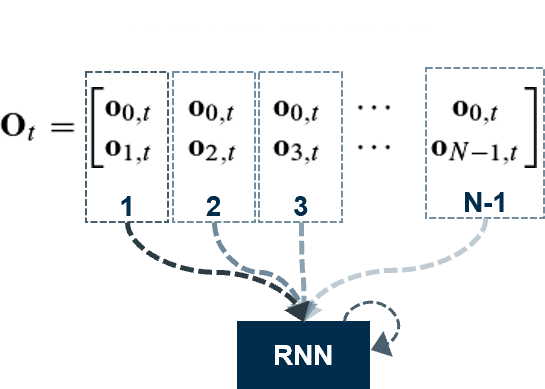}
        \caption{\ac{RNN}}
        \label{fig:ANN-extractor}
    \end{subfigure}
    \hfill
    \begin{subfigure}[b]{0.48\linewidth}
        \centering
        \includegraphics[width=0.95\linewidth]{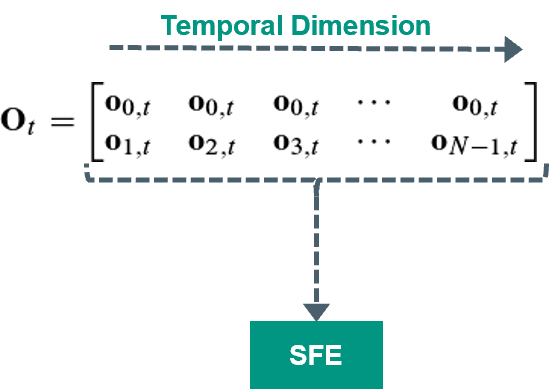}
        \caption{\ac{SFE}}
        \label{fig:SNN-extractor}
    \end{subfigure}
    \vspace{-1em}
    \caption{Comparison between sequence processing of a \ac{RNN} feature extractor approach (a) and the simplified input to our \ac{SFE} approach (b).}
    \label{fig:ANN-SNN-fe-matrix-comparison}
\end{figure}

\subsection{Neuromorphic Encoding}
To realize the neuromorphic part of the actor, conventional \ac{ANN} layers are transformed into their spiking equivalents. This transformation enables the network to process information through discrete spike events rather than continuous activations, allowing temporal dynamics to emerge from the neuron model.
Information transfer between the floating-point and spiking domains requires an encoding and decoding mechanism.
We employ current injection as the encoding scheme, where each floating-point input value is interpreted as a synaptic input current.
This allows for the direct injection of continuous observations without expanding the temporal dimension for neuron size. 
The neuron's membrane potential accumulates this input over time, producing a spike train that represents the encoded signal. 
For decoding, the mean firing rate or the average membrane potential across the temporal dimension is used to recover a continuous representation.

Utilizing a feature extractor before the actor and critic networks enables the use of multi-agent observations and addresses the incompatibility between variable-sized input vectors and the fixed-size input layers of neural networks.
The proposed \ac{SFE} differs from \ac{ANN} approaches in how the information is processed.
We leverage the inherent temporal dimension of \acsp{SNN} and do not rely on an artificial temporal dimension as in \acsp{RNN}.
Given an arbitrarily long input sequence $\mathbf{o}_t$, an \ac{RNN} produces a feature vector $\mathbf{h}_t$ that represents a weighted combination of the observations in the sequence.
Contrary, neuromorphic \ac{SNN}-based approaches introduce an additional inherent temporal dimension.
With our proposed \ac{SFE} (Fig.~\ref{fig:SNN-extractor}), we use this additional temporal dimension by inserting the complete observation matrix $\mathbf{o}_t$ with the agents stacked in the temporal dimension with:
\begin{equation}
\mathbf{O}_t^{H \times T} = \begin{bmatrix} 
\mathbf{o}_{0,t} & \mathbf{o}_{0,t} & \cdots & \mathbf{o}_{0,t} \\
\mathbf{o}_{1,t} & \mathbf{o}_{2,t} & \cdots & \mathbf{o}_{N-1,t}
\end{bmatrix}\mathrm{.}    
\end{equation}
with the robot observation $\mathbf{o}_{0,t}$ and the human observation $\mathbf{o}_{i,t}$ at time step $t$.
For a detailed description of the observation, refer to (Appx.~\ref{appendix:obs}).
The human agents are ordered from closest to farthest away $\mathbf{o}_{N-1,t}$.
By stacking the robots in the temporal dimension of the \ac{SNN}, we achieve a fixed-sized input $H$ while allowing the observation of an arbitrary number of agents.
Using the inherent temporal dimension, we simplified the input for the feature extractor from inserting a sequence of agents to a single matrix.
Additionally, unlike the \ac{RNN} feature extractor, a compression of the output to a feature vector $\mathbf{h}_t$ can be avoided for the actor network, as our approach utilizes a \ac{SAN} that also features a temporal dimension.
This allows outputting the extracted features as a temporally dependent matrix $\mathbf{H}_t$.
With the smaller compression of information, we expect a performance increase in navigation and social compliance.

The observation matrix is additionally sorted so that the closest robot has the highest influence on the output of the \ac{SFE}.
We invert the sequence order from \cite{everett_collision_2021} for our concept, because of the different working principle of an \ac{SFE} compared to an \ac{RNN}.
The first time bin of the input of a \ac{SFE} enters the neuron model with its dynamics and gets outputted at the second time bin.
With the second time bin of the input now entering the neuron dynamic, the information from the first input time bin remains encoded into the neuron dynamic.
Therefore, the output of the third-time bin results is based on the information from the first and second input time bins. 
This illustrates how the initial information of the sequence of an \ac{SFE} influences all future inputs and, therefore, most significantly impacts the accumulated temporal output.

\begin{table}[tb]
\centering
\caption{Assumed Energy (J) per Operation}
\vspace{-0.5em}
\setlength{\tabcolsep}{1pt}       
\label{tab:energy_params}
\begin{adjustbox}{width=\linewidth}
\begin{tabular}{|l|ccc|}
\hline
Device & {Synop Energy} & {Neurop Energy} & Source \\
\hline
CPU x86 
& $\num{8.60e-09}$ & $\num{8.60e-09}$ & \cite{degnan_assessing_2016} \\
CPU ARM 
& $\num{9.00e-10}$ & $\num{9.00e-10}$ & \cite{degnan_assessing_2016} \\
GPU 
& $\num{3.00e-10}$ & $\num{3.00e-10}$ & \cite{degnan_assessing_2016} \\
\hline
SpiNNaker             & $\num{1.33e-08}$ & $\num{2.60e-08}$ & \cite{hoppner_dynamic_2019} \\
SpiNNaker 2           & $\num{4.50e-10}$ & $\num{2.19e-09}$ & \cite{hoppner_dynamic_2019} \\
Loihi                 & $\num{2.71e-11}$ & $\num{8.10e-11}$ & \cite{davies_loihi_2018} \\
\hline
\end{tabular}
\end{adjustbox}
\end{table}

\subsection{Energy Calculation}
The calculation of the energy consumption is inspired by the Nengo framework from \cite{bekolay_nengo_2014}. It estimates the energy per inference by taking synaptic operations as well as neuron updates into consideration:
$E = N_{\mathrm{synops}} \times E_{\mathrm{synop}} + N_{\mathrm{neuron updates}} \times E_{\mathrm{neuron}}$
with $E_{\mathrm{synop}}$ representing the energy necessary per synaptic operation, $N_{\mathrm{synops}}$ representing the number of synaptic operation, $E_{\mathrm{neuron}}$ representing the energy per neuron update, and $N_{\mathrm{neuron updates}}$ is the number of neuron updates and is calculated by multiplying the number of time steps per inference with the number of neurons:
$N_{\mathrm{neuron updates}}=N_{\mathrm{timestep}} \times N_{\mathrm{neurons}}$
Hence, the assumption is made that each neuron is updated at every timestep.
A distinction between neuromorphic and non-neuromorphic hardware is necessary for calculating the synaptic operations.

Neuromorphic hardware is event-driven, synaptic energy is only consumed if an event occurs.
This results in only multiplying every neuron that spikes, represented by the spikestate $s_{t, l, n}\in [1=\mathrm{spiking, }0=\mathrm{else}]$, for all layers $l$ and for each time step $t$ times the number of non-zero synaptic connections between the layers $c_{l,n}$.
This formula can be described as:
\begin{equation}
    N_{\mathrm{synops, neuromorphic}} = \sum_{t,l,n} s_{t,l,n} \times c_{l,n}
\end{equation}

Non-neuromorphic hardware is not event-driven, the assumption is made that all the synaptic connections are computed at each timestep.
Therefore, the total \ac{SYNOPS} is calculated by multiplying the number of synaptic connections times the time steps per inference, regardless of whether they are spiking or not:
\begin{equation}
N_{\mathrm{synops, non-neuromorph}} = t \times \sum_{l,n} c_{l,n}
\end{equation}
%
The hardware-specific Energy per Operation value is displayed in Table~\ref{tab:energy_params}.

\begin{table}[tb]
    \centering
    \caption{PPO Hyperparameters and Rewards }
    \vspace{-0.5em}
    \label{tab:experiment_parameters}
    \begin{tabular}{|l c c || l c|}
        \hline
        Hyperparameter & SD & CUBA & Reward & Value \\
        \hline
        optimizer & \texttt{Adam} & \texttt{Adam} & $R_g$ &  4            \\
        learning rate & \num{2e-4} & \num{9e-5}   & $R_c$   &4  \\
        environment steps & 256 & 128 & $R_{\mathrm{time}}$ &4   \\
        clip range & 0.1 & 0.1 & $R_{gd,1}$ & 0.1 \\
        number of epochs & 4 & 4 & $R_{gd,2}$ & 0.2   \\
        batch size & 256 & 256 & $R_v$ & 0.058  \\
        discount factor $\gamma$ & 0.99 & 0.99 &  $R_{\mathrm{prox}}$& 1.1 \\
        \hline
    \end{tabular}
\end{table}

\section{Evaluation}

\begin{table*}[!ht]
\centering
\caption{Evaluation results with 200 episodes per scenario and seed. Goal, Collision (Col.), \acf{TO}, \acf{PV}, \acf{DR}: ↓}
\vspace{-0.5em}
\label{tab:quantitativeEval}
\setlength{\tabcolsep}{1pt}       
\begin{adjustbox}{width=\textwidth}
\begin{tabular}{|c|c||c|c|c|c|c||c|c|c|c|c||c|c|c|c|c|}
\hline
\multirow{2}{*}{\textbf{Training Scenario}} & \multirow{2}{*}{\textbf{Approach}} &
\multicolumn{5}{c||}{\textbf{Eval. Scenario: Circle Interaction}} &
\multicolumn{5}{c||}{\textbf{Eval. Scenario: Circle Crossing}} &
\multicolumn{5}{c|}{\textbf{Eval. Scenario: Random}} \\
\cline{3-17}
& & Goal (\%) & Col. (\%) & TO (\%) & PV & DR ($\mu \pm \sigma$) &
Goal (\%) & Col. (\%) & TO (\%) & PV & DR ($\mu \pm \sigma$) &
Goal (\%) & Col. (\%) & TO (\%) & PV & DR ($\mu \pm \sigma$) \\
\hline
\hline
\multirow{4}{*}{\textbf{Circle Interaction}} 
& SARL (Baseline) & 93.00 & {0.00} &  7.00   & 93   & 1.08 $\pm$ 0.12  &  \textbf{96.40}   & {0.10}   & 3.50  & 167    &  1.17 $\pm$ 1.07     &  52.50  &   {1.30}   &  46.20 &   191  &    1.67 $\pm$ 1.09   \\
& SI-PPO (Baseline) & \textbf{99.80} & 0.20 & {0.00} & \textbf{7} & {\textbf{1.02 $\pm$ 0.02}} & 93.30 & 4.37 & 2.33 & 169 & {1.06 $\pm$ 0.13} & 76.80 & 4.70 & 18.50 & \textbf{43} & {\textbf{1.19 $\pm$ 0.40}} \\
& Spiking-PPO-SD & 99.65 & 0.35 & {0.00} & 8 & 1.03 $\pm$ 0.03 & 88.85 & 2.50 & 8.65 & \textbf{83} & \textbf{1.05 $\pm$ 0.05} & \textbf{91.00} & 7.25 & {1.75} & 59 & 1.19 $\pm$ 0.49 \\
& Spiking-PPO-CUBA & 98.55 & 1.45 & {0.00} & 46 & 1.08 $\pm$ 0.08 & 93.05 & 5.95 & {1.00} & 131 & 1.12 $\pm$ 0.17 & 58.70 & 17.25 & 24.05 & 80 & 2.17 $\pm$ 4.19 \\
\hline
\hline

\multirow{4}{*}{\textbf{Circle Crossing}} 
& SARL (Baseline) &     39.90    &  {0.4}    &   59.70   &  113  & 2.29 $\pm$ 3.06   & 98.40  &   {0.10}    &   1.5   & 203    &  1.09 $\pm$ 0.27  &   31.30  &   {1.5}  & 67.2  &  218   &   2.26 $\pm$ 2.90              \\
& SI-PPO (Baseline) & 49.57 & 12.20 & 38.23 & \textbf{101} & 2.27 $\pm$ 2.29 & 86.10 & 7.07 & 6.83 & 52 & 1.74 $\pm$ 2.88 & 58.90 & 7.83 & 33.27 & 90 & 6.38 $\pm$ 29.49 \\
& Spiking-PPO-SD & \textbf{95.25} & 4.75 & {0.00} & 115 & \textbf{1.03 $\pm$ 0.04} & \textbf{99.25} & 0.75 & {0.00} & \textbf{27} & {\textbf{1.03 $\pm$ 0.04}} & \textbf{90.95} & 4.95 & {4.10} & \textbf{42} & {\textbf{1.25 $\pm$ 0.67}} \\
& Spiking-PPO-CUBA & 74.10 & 21.90 & 4.00 & 223 & 1.18 $\pm$ 0.29 & 98.30 & 1.70 & {0.00} & 41 & 1.10 $\pm$ 0.10 & 62.10 & 15.00 & 22.90 & 71 & 2.41 $\pm$ 8.79 \\
\hline
\hline

\multirow{4}{*}{\textbf{Random}} 
& SARL (Baseline) &   \textbf{98.40}    &   {0.00}   &     1.6   &   100    &   1.09 $\pm$ 0.10  &  99.40  &   {0.10}   &  0.50  &  172  &   1.12 $\pm$ 0.15    &  \textbf{96.40}  &   {0.20}  &  {3.4}   &  72   &   \textbf{1.10 $\pm$ 0.13}     \\
& SI-PPO (Baseline) & 68.03 & 5.60 & 26.37 & 159 & 2.56 $\pm$ 6.03 & 68.40 & 8.60 & 23.00 & 179 & 2.50 $\pm$ 5.71 & 80.40 & {2.37} & 17.23 & 66 & {2.40 $\pm$ 5.69} \\
& Spiking-PPO-SD & 98.35 & 1.60 & {0.05} & \textbf{63} & {\textbf{1.02 $\pm$ 0.06}} & \textbf{99.50} & {0.50} & {0.00} & \textbf{30} & {\textbf{1.02 $\pm$ 0.02}} & 88.55 & 4.45 & 7.00 & \textbf{40} & 1.27 $\pm$ 0.75 \\
& Spiking-PPO-CUBA & 88.90 & 3.95 & 7.15 & 111 & {1.28 $\pm$ 0.73} & 94.65 & 2.85 & 2.50 & 71 & 1.16 $\pm$ 0.42 & 58.50 & 15.90 & 25.60 & 69 & 2.20 $\pm$ 2.55 \\
\hline
\end{tabular}
\end{adjustbox}
\end{table*}

We adhere to the principles and guidelines for evaluating social navigation, assessing the algorithm’s behavior in response to dynamic obstacles within a reproducible simulation as a preliminary stage before real-world experiments \citep{FrancisPrinciplesandGuidelinesforEvaluating2023}.
First, we compare the training and generalization performance on different scenarios for two different \ac{SNN} neuron types. 
Subsequently, we compare the quantitative social navigation results with the socially integrated SI-PPO \citep{FloegelSociallyIntegratedNavigation2024} and the socially aware SARL \citep{ChanganChenCrowdRobotInteraction:CrowdawareRob2019} policies, and provide a qualitative analysis of the learned social behavior.
Finally, we evaluate the estimated power consumption of our hybrid approach on neuromorphic hardware.

\subsection{Experimental Setup}

For training and evaluation, we utilize the microscopic social-physiological environment described in \cite{FloegelSociallyIntegratedNavigation2024}.
All \ac{DRL} policies are trained with \textit{Stable Baselines3} \citep{AntoninRaffinStableBaselines3ReliableReinforcemen2021}.
For our proposed approach, the \ac{PPO} hyperparameters are obtained through a hyperparameter search performed with \textit{Optuna} \citep{optuna_2019} and are stated in Table~\ref{tab:experiment_parameters}.
Other key parameters include the neuron-parameter for \ac{SD} \ac{SFE}: $V_{th}=0.14$, $\tau_{grad}=0.82$, $s_{grad}=0.16$; the \ac{SD} \ac{SAN}: $V_{th}=0.26$, $\tau_{grad}=0.44$, $s_{grad}=0.58$; the \ac{CUBA} \ac{SFE}: $V_{th}=0.31$, $\alpha_I=0.77$, $\alpha_V=0.49$, $\tau_{grad}=0.55$, $s_{grad}=2.34$; and the \ac{CUBA} \ac{SAN}: $V_{th}=0.93$, $\alpha_I=0.16$, $\alpha_V=0.78$, $\tau_{grad}=0.21$, $s_{grad}=3.47$.

To evaluate the performance and compare the different approaches, we consider three scenarios: a circle interaction scenario ($8$ agents), a circle crossing ($8$ agents), and a random scenario ($10$ agents).
For domain randomization, start and goal locations are randomly selected within the red regions shown in Fig.~\ref{fig:training}, while the robot is randomly allocated to one of the blue start-goal pairs.
In addition, the agents’ preferred speeds and the proxemic radii of humans are drawn from $\mathcal{U}[0.5, 1.0]$ and $\mathcal{U}[0.3, 0.7]$, respectively.
Note that the robot cannot directly observe the proxemic radius, only implicitly through interaction behavior, since humans actively maintain their own proxemic radius.
For further insights, refer to \cite{FloegelSociallyIntegratedNavigation2024}.
The training on the circle interaction and circle crossing scenario is conducted over $\num{6e6}$ steps and on the random scenario for $\num{1e7}$ steps due to the higher complexity.
To assess the generalization capability, we cross-validate the trained policies by evaluating them on the other two scenarios, thereby assessing the agent's adaptability and the impact of the different training environments.
Since the SARL baseline employs a one-step lookahead, its training and evaluation are computationally more demanding.
Therefore, SARL is trained with $5$ random seeds \footnote{We will provide the results across $10$ seeds for the final submission}, whereas all other policies are trained with $10$ seeds. 
All policies are evaluated for $200$ episodes per converging seed and per scenario.

\subsection{Results}
\subsubsection{Quantitative Evaluation:}
\begin{figure}[tb]
    \vspace{-0.5em}
    \centering
    \includegraphics[width=0.95\linewidth]{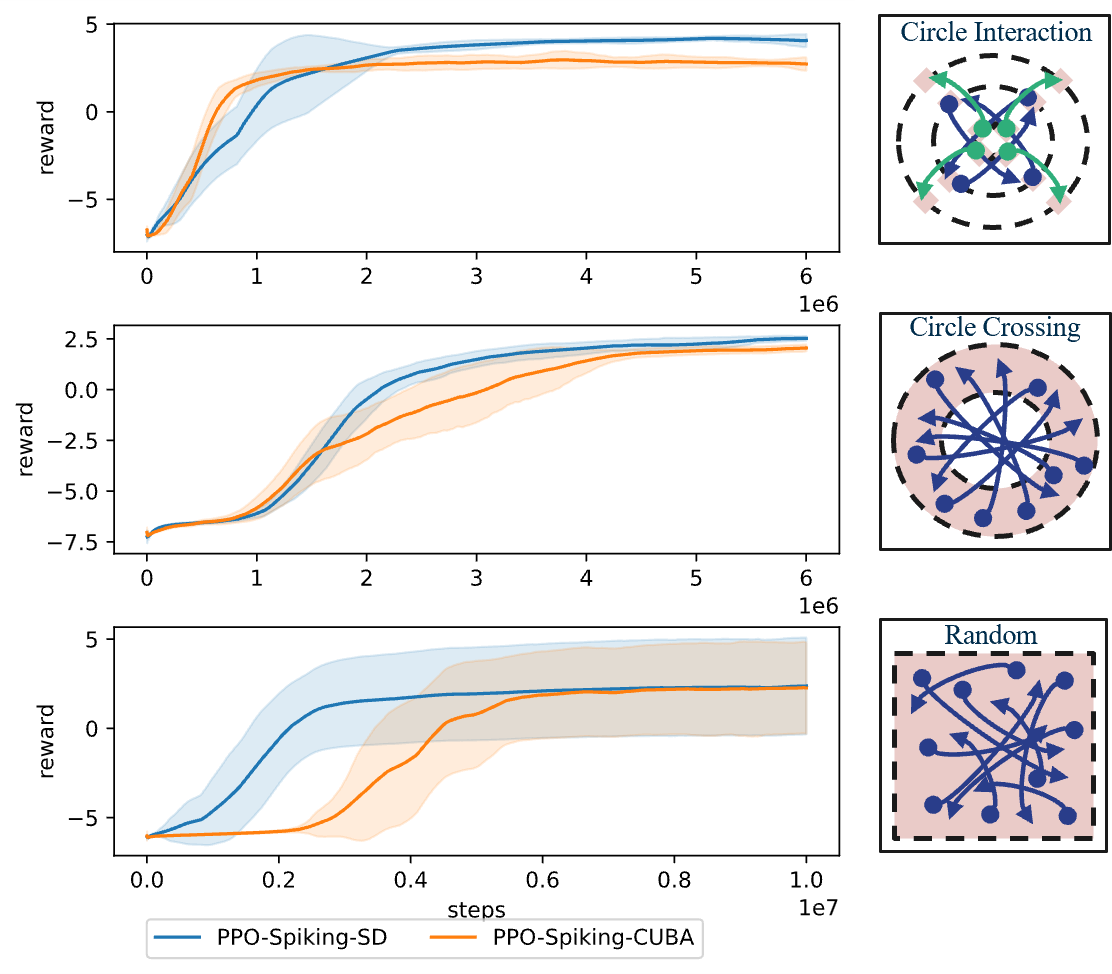}
    \vspace{-1em}
    \caption{Training rewards of our approach with \ac{SD} and \ac{CUBA} neurons across $10$ seeds and different scenarios. The start and goal positions are sampled in red areas.}
    \label{fig:training}
\end{figure}

The training rewards in Fig. \ref{fig:training} illustrate stable and converging training of our hybrid method across all three scenarios for both methods. 
Since the circle interaction scenario has only slight randomization, the training converges the fastest.
Circle crossing introduces more randomness due to the larger spawn areas for the start and goal positions, leading to later convergence.
The effect is particularly pronounced when trained on the random scenario, which has the highest level of randomization.
Besides the randomization regarding start and goal position, it also introduces a higher variance in the distance the agents need to travel and less interaction between agents. 
The proposed approach with the \ac{SD} neurons achieves higher training rewards, which is especially prominent in the circle interaction and circle crossing scenario.
Additionally, Spiking-PPO-SD demonstrates fast and stable training across all scenarios and seeds. 
Our experiments revealed that Spiking-PPO-CUBA is highly sensitive to hyperparameters, resulting in less stable and slower convergence, whereas Spiking-PPO-SD provides a robust, hyperparameter-insensitive alternative.

We consider the percentage of the robot reaching the goal, colliding with a human, the rate at which the robot runs into a \acf{TO}, the number of \acf{PV}, and the \acf{DR} to evaluate the navigation performance.
The \ac{DR} is the fraction of traveled distance and ideal distance, which is a straight line to the goal at the preferred velocity. 
The evaluation results in Table~\ref{tab:quantitativeEval} reveal that both proposed approaches show competitive or superior performance in all scenarios.
When trained on the circle crossing scenario, the Spiking-PPO-SD outperforms both baselines and exhibits superior performance compared to the Spiking-PPO-CUBA approach. 
Across all scenarios, the Spiking-PPO-SD exhibits higher generalization capabilities, as evidenced by a higher success rate in non-training scenarios. 
Even better than the SARL baseline, which includes an imitation learning phase with ORCA expert trajectories to initialize the networks for \ac{DRL} training. 
Across all scenarios, the socially integrated navigation policies exhibit the lowest proxemic violations, indicating an individual consideration of human interaction behavior. 

Concluding the results in Table~\ref{tab:quantitativeEval}, the Spiking-PPO-SD demonstrates superior robustness, generalization, and adaptation to human behavior across all scenarios.
It has the lowest \ac{PV} and \ac{DR}, indicating the highest social integration in terms of adapting to individual human behavior.
Therefore, we will only further evaluate the Spiking-PPO-SD approach.
We suspect that the hyperparameter sensitivity of the CUBA neurons contributes to the poorer performance, and CUBA neurons require more time steps to encode information into an information-rich spike train, thereby reducing their hyperparameter sensitivity.
This also explains the superior performance of Spiking-PPO-SD, since the \ac{SD} neurons can encode information at a single timestep through graded spikes.
\begin{figure}[tb]
    \centering
    \includegraphics[width=0.9\linewidth]{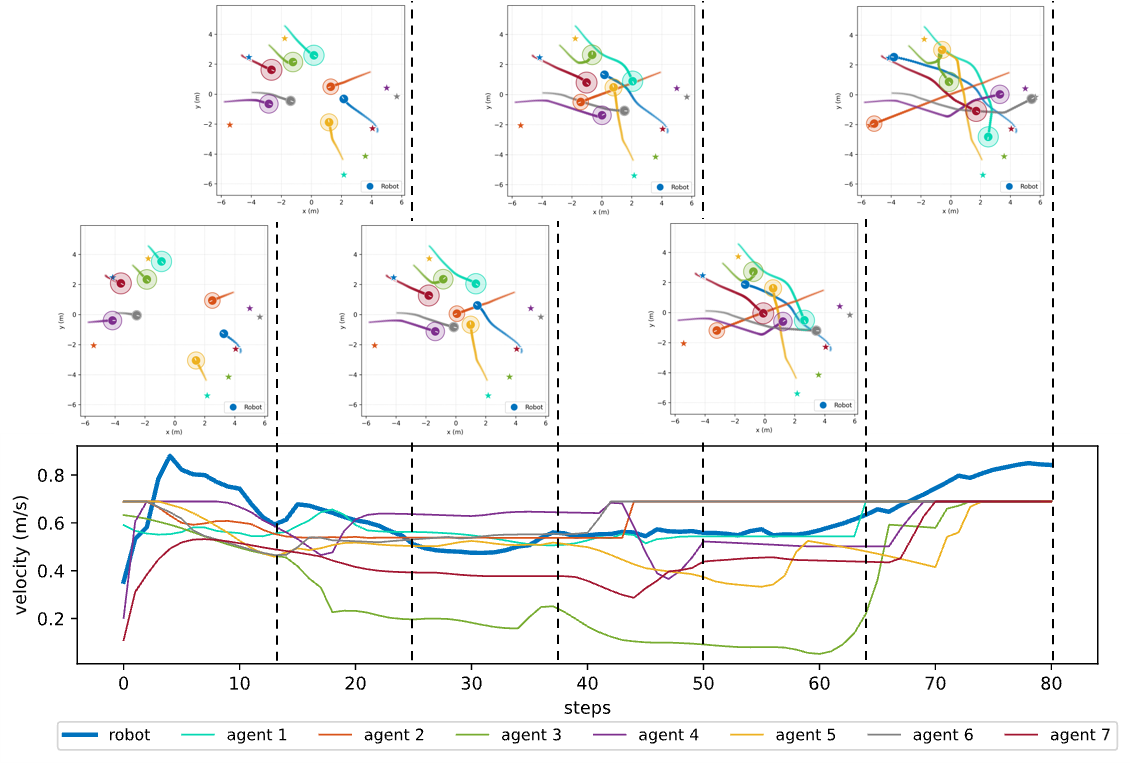}
    \vspace{-0.5em}
    \caption{Socially Integrated: The Spiking-PPO-SD policy is adaptive and respects individual human proxemics, leading to smooth trajectories for all agents.}
    \label{fig:behaviorCC}
\end{figure}
\begin{figure}
    \centering
    \includegraphics[width=0.9\linewidth]{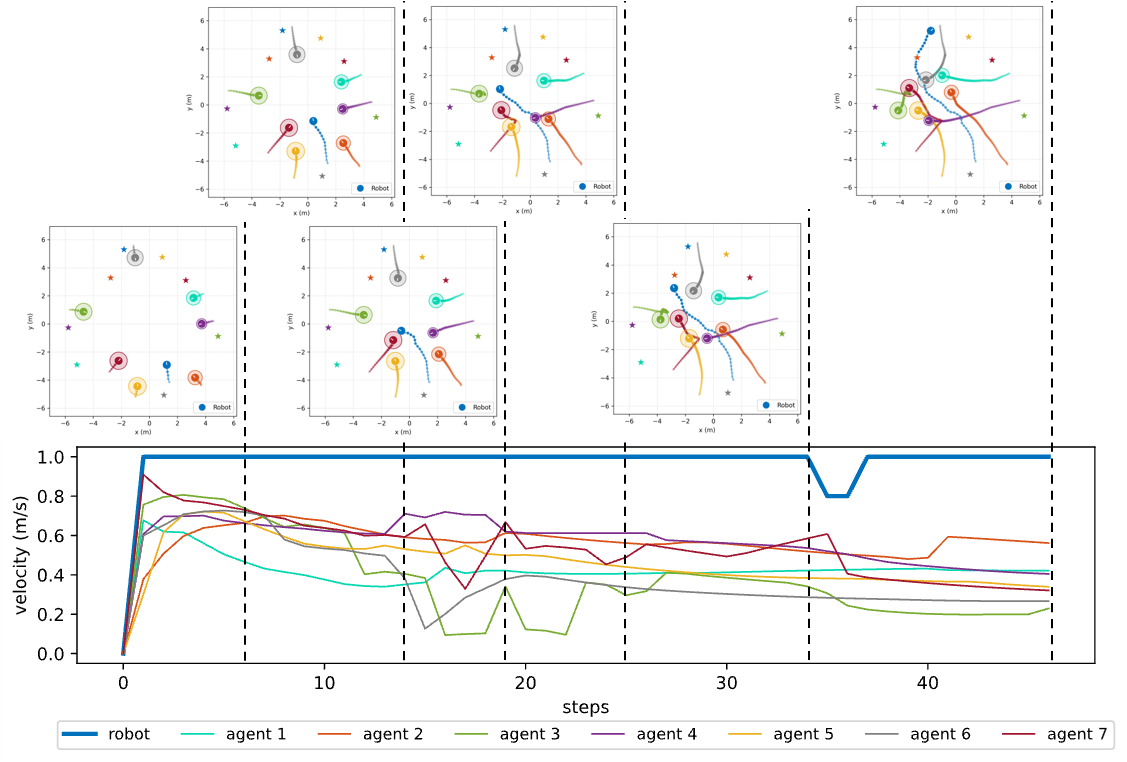}
    \vspace{-0.5em}
    \caption{Socially Aware: the SARL policy focuses on ego navigation, leading to a negative impact on human agents' trajectories. }
    \label{fig:behaviorSARL}
\end{figure}

\subsubsection{Qualitative Evaluation:}
As a qualitative metric, we evaluate the robot's trajectory and speed, as well as its interaction behavior with human agents.
The ideal behavior is considerate, robust, and efficient, characterized by socially integrated navigation, which enables the robot to adapt to and coordinate with the behavior of surrounding agents, resulting in efficient navigation for all agents involved.
The Spiking-PPO-SD in Fig.~\ref{fig:behaviorCC} exhibits socially integrated behavior by respecting each agent's proxemic radius and adjusting its speed accordingly to that of surrounding agents.
The robot navigates closely past the agents, adjusting its heading and slowing as needed while still following a goal-driven trajectory.
Another learned behavior that shows social integration is passing behind another agent.
At step $25$ in Fig.~\ref{fig:behaviorCC}, the robot slows down, adjusts its heading to pass behind the second agent at step $38$, and accelerates at step $64$ once the social interaction is resolved.
In contrast, the socially aware SARL policy in Fig.~\ref{fig:behaviorSARL} mainly navigates at its maximum speed and does not adapt to human agents' behavior. 
The robot reaches the center point first, but then pushes other agents away as depicted in steps $19$ and $25$, and violates the proxemic radius of the red agent in step $19$. 
This leads to non-smooth trajectories of human agents in the surroundings and a negative impact on the crowd behavior. 
These results reveal that the ego-centric perspective of socially aware navigation policies may lead to goal-directed robot trajectories but has a negative impact on the behavior of other agents. 
In contrast, the socially integrated navigation policy with its adaptive behavior leads to goal-driven trajectories of all agents involved and even to short traveled distances as revealed through the \ac{DR} in Table.\ref{tab:quantitativeEval}.

\subsubsection{Energy Consumption:}

\begin{figure}
    \centering
    \includegraphics[width=0.9\linewidth]{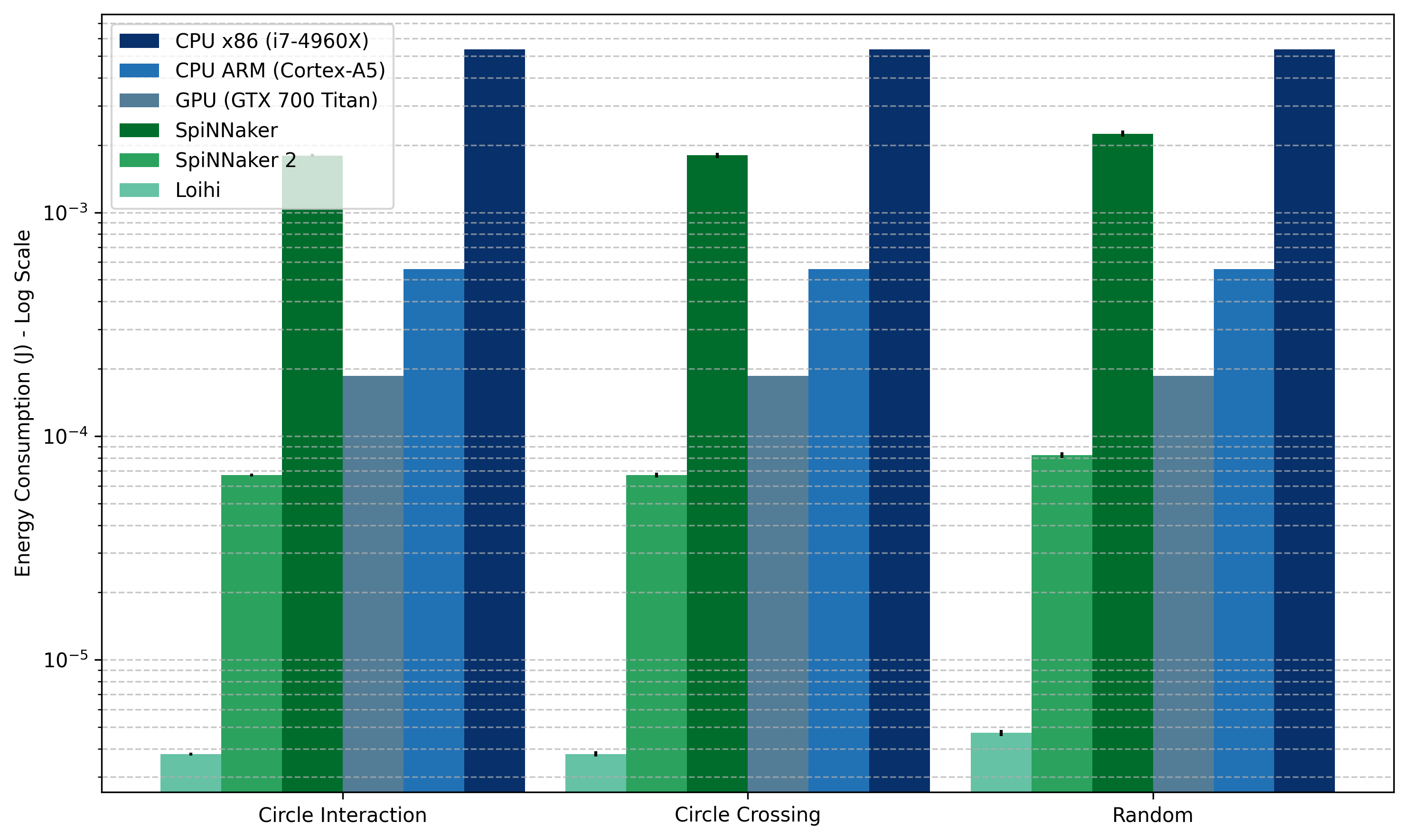}
    \caption{Estimated energy consumption of the Spiking-PPO-SD on different hardware and scenarios.}
    \label{fig:energyConsumption}
\end{figure}

The estimated energy consumption of the Spiking-PPO-SD over $200$ evaluation episodes across all three scenarios is displayed in Fig.~\ref{fig:energyConsumption}.
Since sparsity affects only the neuromorphic estimated energy consumption during synaptic updates, only the neuromorphic hardware devices vary across evaluation samples. In contrast, the conventional hardware produces the same estimated energy consumption across evaluation samples.
The estimated energy consumption is the lowest for the Loihi with $3.79 \pm 0.06 \mu J$ on the circle interaction, $3.79 \pm 0.10 \mu J$ on the circle crossing, and $4.71 \pm 0.14 \mu J$ on the random scenario.
The second lowest energy consumption was estimated on the SpiNNaker 2 with $66.86 \pm 1.00 \mu J$ on the circle interaction, $66.93 \pm 1.68 \mu J$ on the circle crossing, and $82.17 \pm 2.38 \mu J$ on the random scenario.
The first conventional hardware is the GPU (GTX 700 Titan) with $186.33 \mu J$ followed by the ARM CPU (Cortex-A5), of $559.00 \mu J$.
Finally, the SpiNNaker is estimated to consume $1.19 \pm 0.03 mJ$ on the circle interaction, $1.15 \pm 0.03 mJ$ on the circle crossing, and $1.29 \pm 0.08 mJ$ on the random scenario, while the highest estimated energy consumption was reached by the x86 CPU (i7-4960X) with $5.34 mJ$.
The $1.60-1.69$ magnitude lower estimated energy consumption obtained by running our Spiking-PPO-SD on 
a neuromorphic device, such as the Loihi, compared to running on conventional hardware, showcases the 
potential energy improvements that can be obtained through neuromorphic computing.

\section{Conclusion}
This work presents SINRL, a hybrid \acf{DRL} approach for socially integrated navigation in pedestrian-rich environments using \acf{SNN}.
We propose a novel \acf{SFE} that enables the efficient encoding of multi-agent observations with the combination of a \acf{SAN}, and a \acf{ANN} critic for stable training.
The \ac{SFE} efficiently encodes social-temporal interactions between the crowd and humans. 
Our experiments reveal that \acf{SD} neurons provide more stability during training and yield better results compared to \ac{CUBA} neurons. 
The proposed navigation approach outperforms existing methods in social adaptation and navigation capabilities.
In addition, the neuromorphic implementation has potentially lower energy efficiency compared to a conventional \ac{ANN} approach.
Future work aims to validate the potential energy benefits by deploying the trained \acp{SNN} on the Loihi 2.


\bibliography{indices/references}             
                                                   
\appendix

\section{Observation}\label{appendix:obs}    
We follow the proposed observation in \cite{FloegelSociallyIntegratedNavigation2024}. 
The observation is divided into the robot $\bm{o}_{0,t} =  [d_g, \Delta, \bm{p}_g, \theta,  v_{\text{pref}}, r_0]$ and human observation $\bm{o}_{i,t} = [\bm{\Bar{o}}_i, \bm{\hat{o}}_{i,t}] $. 
Based on humans' observable states, the robot observes the human $i$ at time $t$ with $\bm{\Bar{o}}_i =  [r_i,  r_i + r_0]$ and $\bm{\hat{o}}_{i,t} =  [d_t,  \Delta, \bm{p}_t,  \Delta, \bm{v}_t]$.
The robot observes its state based on the relative position to the goal $\Delta \bm{p}_g = \bm{p}_g - \bm{p}_t$, the direct distance to the goal $d_g = ||\Delta \bm{p}_g||_2$, the heading $\theta$, the preferred velocity $v_{\text{pref}}$, and the personal radius $r$.
We distinguish the human observation into a constant part $\bm{\Bar{o}_i}$, which remains unchanged over time, and a time-varying part $\bm{\hat{o}}_{i,t}$, which varies throughout the steps. 
The constant part contains the human radius $r_i$ and the combined radius $r_i + r_0$. 
The temporal observation $\bm{\hat{o}}_{i,t}$ contains the relative position $\Delta \bm{p}_t =\bm{p}_{0,t} - \bm{p}_{i,t}$, direct distance $d_t = ||\Delta \bm{p}_t||_2$, and relative velocity $\Delta \bm{v}_t = \bm{v}_{0.t} - \bm{v}_{i,t}$ between robot and human. 
To consider the partial history, we concatenate the last $k$ temporal observations to the aggregated observation $\bm{o}_{i,t}$ of human $i$ at timestep $t$
\begin{align}
    \bm{o}_{i,t} = 
    \begin{bmatrix}
        \bm{\Bar{o}}_{i,t}& \bm{\hat{o}}_{i,t}& \bm{\hat{o}}_{i,t-1} & \cdots& \bm{\hat{o}}_{i,t-k} 
    \end{bmatrix} .
\end{align}

\section{Reward System}\label{appendix:rew}
In the socially integrated reward formulation from \cite{FloegelSociallyIntegratedNavigation2024}, each agent in the environment has a reward system. 
Humans reward the robot based on the interaction between the human and the robot.
From this principle, the robot accumulates all human rewards with
\begin{align}
    R_{\text{SA}} = \frac{1}{M} \sum_{i=1}^M \lambda_i R_i
\end{align}
at each time step.
With the reward $R_i$ of human $i$, a scaling factor $\lambda_i$, and the number of humans $M$ within the social integration radius $r_{\text{SI}}$ around the center of the robot. 
If the human $i$ is within the social integration radius of the robot $d_{0i} < r_{\text{SI}}$, the human reward is given at every time step by
$R_i =  - R_v \cdot |v_{i,t} -v_{0,t}| -  R_{\text{prox}}$.
The reward comprises velocity deviations scaled with $R_v$ and rewards violations of humans' personal space $d_{i0} < r_{i,\text{prox}}$ with $R_{\text{prox}}$.
In addition, the robot is rewarded for efficient navigation with $R_{\text{Nav}}$, which leads to the total reward  $R_t = R_{\text{Nav}} + R_{\text{SA}}$ at time step $t$.
The navigation reward teaches efficient navigation toward the goal while avoiding collisions.
\begin{equation}
R_{\text{Nav}} = \begin{cases}
    + R_g & \textrm{if} \hspace{3mm} \bm{p}_t=\bm{p_g} \\
    - R_c & \textrm{if} \hspace{3mm} d_{0i} \leq 0 \\
    - R_{\text{time}} & \textrm{if} \hspace{3mm} \text{timeout} \\
    + R_{gd,1} \cdot |\Delta d_g|  & \textrm{if} \hspace{3mm} \Delta d_g  > 0 \\
    - R_{gd,2} \cdot  |\Delta d_g|  & \textrm{if} \hspace{3mm} \Delta d_g  < 0 \\
    \end{cases}
\end{equation}
The robot is encouraged to take steps toward the goal with $R_{gd,1}$, where $\Delta d_g = d_{g,t} - d_{g,t-1}$, and reach the goal with $R_g$. 
Conversely, the robot is penalized for collisions with other agents with $R_c$, running into timeouts with $R_{\text{time}}$, and taking steps away from the goal with $R_{gd,2}$.

\end{document}